\documentclass{edm_template}

\usepackage{graphicx}
\graphicspath{{images/}}

\usepackage{booktabs} % For professional looking tables
\usepackage{multirow}
\usepackage{adjustbox}
\usepackage{tabularx}
\usepackage{multicol}
\usepackage{csvsimple}
\usepackage{pgfplotstable}
\newcolumntype{L}[1]{>{\raggedright\arraybackslash}p{#1}}
\usepackage{makecell}

\usepackage{url}

\usepackage{float}

\usepackage{amsmath, amssymb, bm}
\usepackage{amssymb}
\usepackage{amsmath}
\usepackage{bm}
\usepackage[mathcal]{eucal}

\def \a {\mathbf{a}}
\def \b {\mathbf{b}}

\def \e {\mathbf{e}}
\def \f {\mathbf{f}}

\def \h {\mathbf{h}}

\def \k {\mathbf{k}}

\def \q {\mathbf{q}}
\def \r {\mathbf{r}}

\def \v {\mathbf{v}}
\def \w {\mathbf{w}}
\def \x {\mathbf{x}}
\def \y {\mathbf{y}}

\def \A {\mathbf{A}}
\def \B {\mathbf{B}}

\def \M {\mathbf{M}}

\def \P {\mathbf{P}}

\def \S {\mathbf{S}}

\def \W {\mathbf{W}}
\def \X {\mathbf{X}}

\def \Lcal {\mathcal{L}}

\def \Rbb {\mathbb{R}} % include the self-defined math symbol

\usepackage[numbers]{natbib}
\usepackage{balance}

\usepackage{comment}

\title{Deep-IRT: Make Deep Learning Based Knowledge Tracing Explainable Using Item Response Theory}

\numberofauthors{1}
\author{
    \alignauthor
    Chun-Kit Yeung\\
    \affaddr{Find Solution Ai Limited}\\
    \affaddr{Hong Kong}\\
    \email{chunkit@findsolutiongroup.com}
}
\begin{document}

\maketitle 
\begin{abstract}
Deep learning based knowledge tracing model has been shown to outperform traditional knowledge tracing model without the need for human-engineered features, yet its parameters and representations have long been criticized for not being explainable. In this paper, we propose \emph{Deep-IRT} which is a synthesis of the item response theory (IRT) model and a knowledge tracing model that is based on the deep neural network architecture called dynamic key-value memory network (DKVMN) to make deep learning based knowledge tracing explainable. Specifically, we use the DKVMN model to process the student's learning trajectory and estimate the item difficulty level and the student ability over time. Then, we use the IRT model to estimate the probability that a student will answer an item correctly using the estimated student ability and the item difficulty. Experiments show that the Deep-IRT model retains the performance of the DKVMN model, while it provides a direct psychological interpretation of both students and items.
\end{abstract}


\section{Introduction}
With the advancement of digital technologies, online platforms for intelligent tutoring systems (ITSs) and massive open online courses (MOOCs) are becoming prevalent. These platforms produce massive datasets of student learning trajectories about the \emph{knowledge components} (KCs), where KC is a generic term for concept, skill, exercise, item, etc. The availability of online activity logs of students has accelerated the development of learning analytics and educational data mining tools for predicting the performance and advising the learning of students. Among many topics, knowledge tracing is considered to be important for enhancing personalized learning. Knowledge tracing is the task of modeling student's \emph{knowledge state}, which is a general representation of the mastery level of KCs, e.g., a scalar value representing a student ability level, or a vector representation similar to word embedding~\cite{NIPS2015_Mikolov_Word2Vec}. With the estimated students' knowledge state, teachers or tutors can gain a better understanding of the attainment levels of their students and can tailor the learning materials accordingly. Moreover, students may also take advantage of the learning analytics tools to come up with better learning plans to deal with their weaknesses and maximize their learning efficacy.

Generally, the knowledge tracing task can be formalized as follows: given a sequence of student's historical interactions $\X_t = ( \x_{1}, \x_{2}, ..., \x_{t} )$ up to time $t$ on a particular learning task, it predicts some aspects of his next interaction $\x_{t+1}$. Question-and-answer interactions are the most common type in knowledge tracing, and thus $\x_t$ is usually represented as an ordered pair $(q_t, a_t)$ which constitutes a tag for the question $q_t$ being answered at time $t$ and an answer label $a_t$ indicating whether the question has been answered correctly. In many cases, knowledge tracing seeks to predict the probability that a student will answer a question $q_{t+1}$ correctly given the sequence $\X_t$, i.e., $P(a_{t+1} = 1 | q_{t+1}, \X_t)$. 

Many mathematical and computational models have been developed to solve the knowledge tracing task. These models can be grouped into two categories~\cite{EDM2016_Khajah_How}: 
\begin{enumerate}
    \item a highly structured model whose parameters have a direct meaningful interpretation, e.g., Bayesian knowledge tracing (BKT)~\cite{UMUAI1994_Corbett_BKT} and performance factors analysis (PFA)~\cite{AIED2009_Pavlik_PFA};
    \item a highly complex but general-purpose model whose parameters are difficult to interpret, e.g., deep knowledge tracing (DKT)~\cite{NIPS2015_Piech_DKT} and dynamic key-value memory network (DKVMN) for knowledge tracing~\cite{WWW2017_Zhang_DKVMN}.
\end{enumerate}
The former category typically provides more insight besides the prediction result, while the latter usually performs better without requiring substantial feature engineering by humans. To the best of our knowledge, there has not yet been a model that is highly complex and general-purpose, yet simultaneously explainable. Therefore, it is appealing to devise a model that inherits the merits of these two categories. 

In this paper, we propose \emph{deep item response theory} (Deep-IRT) to make the deep learning based knowledge tracing model explainable. The Deep-IRT model is inspired by the Bayesian deep learning~\cite{IEEE2016_Wang_BDL} and is a synthesis of a deep learning model and a psychometric model. Specifically, the Deep-IRT model utilizes the DKVMN model to process input data and return psychologically meaningful parameters of the IRT model. The DKVMN model performs feature engineering job to extract latent features from student's historical question-and-answer interactions. Then, the extracted latent features are used to infer the difficulty level of and the student ability on each KC over time. Based on the estimated student ability and the KC difficulty level, the IRT model predicts the probability that the student will answer a KC correctly. By formulating the knowledge tracing task with both the DKVMN model and the IRT model, we are getting the merits from these two models. The Deep-IRT model benefits from the advance of deep learning techniques, e.g., capturing features that are hard to be human-engineered. On the other hand, we empower the explainability by introducing a well-known psychometric model which can be easily understood by many people. 

Our experiments show that the proposed Deep-IRT model retains the performance of the DKVMN model. We also conduct analyses on the difficulty level and the student ability learned by the Deep-IRT model. Analyses reveal that the difficulty level estimated by the Deep-IRT model aligns with other traditional methods, e.g., the IRT model and the item analysis~\cite{Washington_ItemAnalysis}. However, the Deep-IRT model still suffers from the reconstruction issue which is discovered in the DKT model~\cite{LS2018_Yeung_DKTP}.

Our main contributions are summarized as follows:
\begin{enumerate}
    \item The proposed Deep-IRT knowledge tracing model is capable of inferring meaningful estimation of student ability and KCs' difficulty level while simultaneously retains the predictive power of the deep learning based knowledge tracing model.
    
    \item The Deep-IRT model potentially provides an alternative way for estimating KC's difficulty level by utilizing the entire learning trajectory, rather than the traditional educational testing environment.
    
    \item We propose to use a deep learning model to output parameters of a psychometric model so as to leverage the deep learning capability and provide explainable psychometric parameters. This idea can be applied elsewhere apart from the knowledge tracing task.
\end{enumerate}
\section{Literature Review}

\begin{figure*}[!h]
	\centering
    \includegraphics[width=0.8\linewidth]{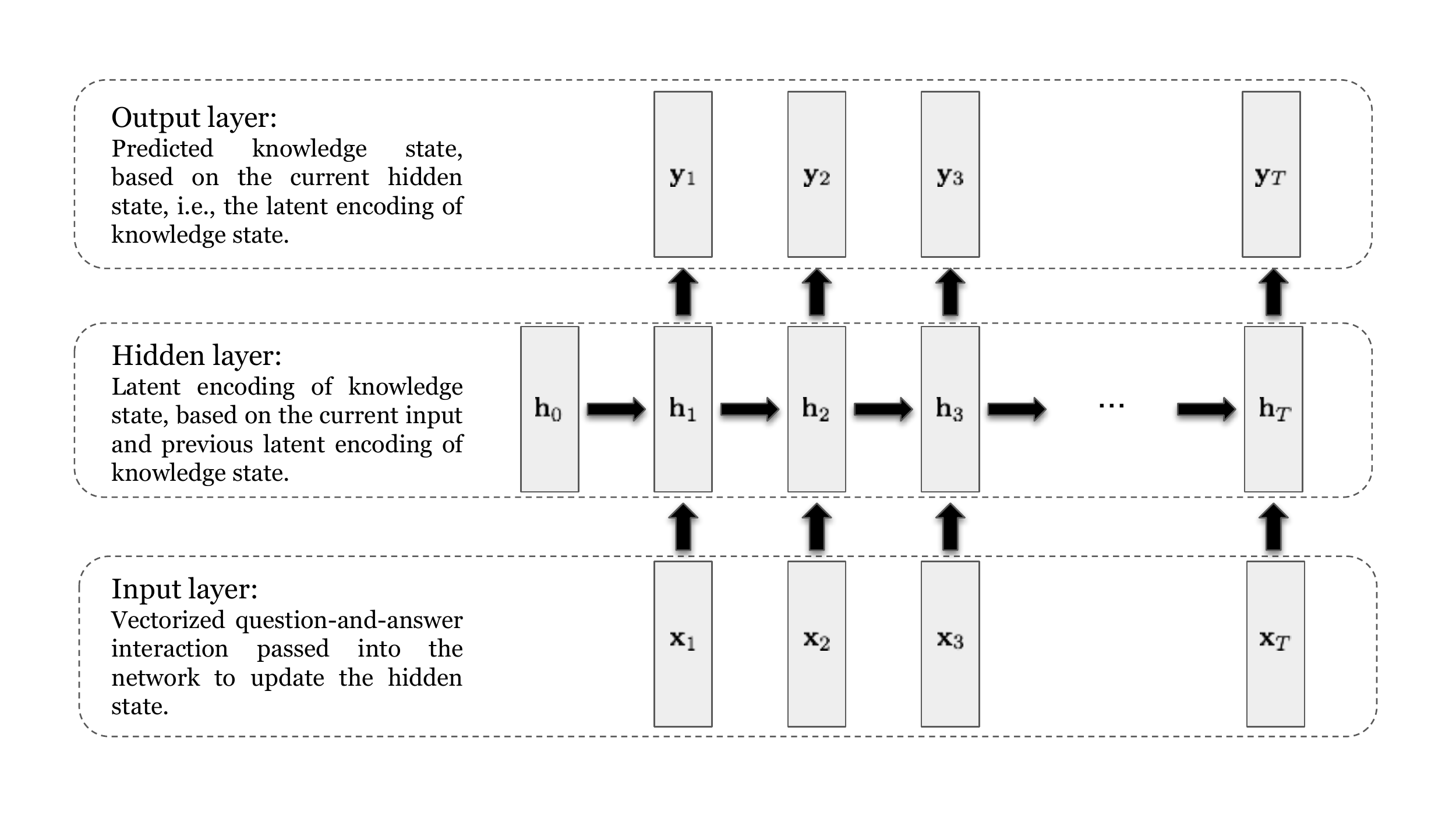}
    \caption{Unfolded version of the RNN architecture for DKT. The hidden state is processed differently in the vanilla RNN or LSTM-RNN. $\h_0$ is the initial hidden state in the RNN, and it is usually initialized randomly or to a zero vector.}
    \label{fig:rnn_architecture}
\end{figure*}

\subsection{Item Response Theory}
Item response theory~\cite{rash1960probabilistic} (IRT) has been used in the educational testing environment since the 1950s. It outputs the probability $P(a)$ that a student will answer an item (i.e., a question) $j$ correctly during a test, based on the student's ability level $\theta$ and the item's difficulty level $\beta_j$ (in the simplest one-parameter IRT\footnote{The one-parameter IRT has the property of specific objectivity. It means that the rank of the item difficulty is the same for all students independent of ability, and the rank of the person ability is the same for all items independent of difficulty.}.) 
This probability is defined by an item response function which has the following characteristics: If a student has a higher ability level, the student has a higher probability to answer an item correctly; on the other hand, if an item is more difficult, a student has a lower probability to answer the item correctly. Most commonly, the logistic regression model is used in the IRT model as the item response function:
\begin{align}
    P(a) 
    &= \sigma \left( \theta - \beta_j \right)
    = \frac{1}{1+\exp(-(\theta - \beta_j))}
\end{align}
where $\sigma(\cdot)$ is the sigmoid function. 

In addition to estimating the probability $P(a)$, the IRT model has also been widely used to estimate the student ability~$\theta$ and the item difficulty level~$\beta_j$. Yet, as the IRT model is originally devised for the educational testing environment, the model assumes that the student's ability is not changing during the test. Thus, it cannot be applied directly to the task of knowledge tracing, where student's knowledge state is changing over time. 

\subsection{Knowledge Tracing}
To tackle the knowledge tracing task, researchers have been investigating mathematical and computational models since the 1990s. Various approaches, ranging from probabilistic models to deep neural networks, have been developed over the past three decades.

\subsubsection{Bayesian Based Knowledge Tracing}
The Bayesian knowledge tracing (BKT) model was proposed by Corbett and Anderson~\cite{UMUAI1994_Corbett_BKT} during the 1990s. It is used to model a skill acquisition process by a hidden Markov model (HMM). However, many simplifying assumptions adopted by the BKT model are unrealistic. One of the assumptions is that all students and questions of a same skill are treated equally in the BKT model. Thus, researchers have investigated in various individualizations on the BKT model. Some researchers have empowered the individualization of the BKT model on both skill-specific parameters~\cite{UMAP2010_Pardos_BKT, UMAP2011_Pardos_KT} and student-specific parameters~\cite{IAIED2013_Yudelson_individualized}. Some other researchers has also investigated the synthesis of the BKT model and the IRT model~\cite{PALE2014_Khajah_Integrating,EDM2016_Wilson_Back} to empower the individualization over questions and students. However, it should be noted that such extensions often require considerable feature engineering efforts and may incur a significant increase in the computational requirements.

\subsubsection{Factors Analysis Based Knowledge Tracing}
In the 2000s, learning factors analysis (LFA)~\cite{ITS2006_Cen_LFA} and performance factors analysis (PFA)~\cite{AIED2009_Pavlik_PFA} are proposed to tackle the knowledge tracing task using the logistic regression model. Both models are similar to the IRT model, yet they estimate the probability that a student will answer a question correctly by learning skill-level parameters. LFA is formulated as follows:
\begin{equation}
P(a) = \sigma ( \theta + \sum_{j \in skills} (\gamma_j N_{j} - \beta_j) ),
\end{equation}
where~$\sigma(\cdot)$ is the sigmoid function,~$\theta$,~$\gamma_j$ and~$\beta_j$ are the model parameters, and~$N_{j}$ is the input to the model. Similar to the IRT model, $\theta$ and $\beta_j$ can be conceived as the ability of a student and the difficulty level of a skill~$j$, respectively. $N_{j}$ means the number of attempts that the student has on the skill~$j$, so~$\gamma_j$ can be interpreted as the learning rate for the skill~$j$. 

After the emergence of the LFA model, Pavlik et al., who believed that student performance has a higher influence than the student ability in tackling the KT task, proposed the PFA model which offers higher sensitivity to student performance rather than student ability~\cite{AIED2009_Pavlik_PFA}. Concretely, it discards the parameter $\theta$ in the LFA model and splits the input $N_{j}$ into $S_{j}$ and $F_{j}$, which represent the number of successful and failed attempts, respectively, the student has on the skill $j$. The PFA model is formulated as follows:
\begin{equation}
P(a) = \sigma ( \sum_{j \in skills} (\alpha_j S_{j} + \rho_j F_{j} - \beta_j) ),
\end{equation}
where $\alpha_j$ and $\rho_j$ are the new model parameters. Similarly, both $\alpha_j$ and $\rho_j$ can be considered as the learning rate for the skill~$j$ when it is applied successfully and unsuccessfully, respectively. In analogue with the IRT model, we can deem that the PFA model treats $\alpha_j S_{j} + \rho_j F_{j}$ to be the student ability $\theta$ on the skill $j$, such that a student can have different ability levels on different skills. It turns out that the PFA model performs better than the LFA model~\cite{JEDM2015_Galyardt_RPFA}.

As we can see from the model formulation, both LFA and PFA models can handle a learning task that is associated with multiple skills. However, it should be noted that they require manual labeling about the skills involved in solving the learning task, and they cannot deal with the inherent dependency among skills. Furthermore, the LFA assumes that $P(a)$ increases monotonically with the total number of attempts, so it is impossible to transit a student's knowledge state from learned to unlearned. Although the PFA model relieves this assumption by introducing the count of failed attempts, it is still difficult for the PFA model to decrease $P(a)$ once a student has numerous successful attempts.
Even this issue is alleviated in the recent-PFA model~\cite{JEDM2015_Galyardt_RPFA} which uses an exponentially weighted average approach to obtain the recent correct rate of a student, we consider that the features used in the LFA and PFA models are still relatively simple and human-engineered, so it cannot adequately represent the students' knowledge state.

\subsubsection{Deep Learning Based Knowledge Tracing}

Recently, with a surge of interest in deep learning models, deep knowledge tracing (DKT)~\cite{NIPS2015_Piech_DKT}, which models student's knowledge state based on a recurrent neural network (RNN), has been shown to outperform the traditional models, e.g.,  BKT and PFA, without the need for human-engineered features such as recency effect, contextualized trial sequence, inter-skill relationship, and students' ability variation~\cite{EDM2016_Khajah_How}. 

% It should also be noted that the DKT model can be applied on both question or skill. Thus, in the following, we will use the generic term knowledge component (KC) to denote skill and question. 

In DKT, an interaction $(q_t, a_t)$ is first transformed into a fixed-length input vector $\x_{{t}}$ using one-hot encoding. After the transformation, DKT passes the $\x_t$ to the hidden layer and computes the hidden state $\h_{{t}}$ using long short-term memory (LSTM) cells~\cite{Neural1997_Hochreiter_LSTM}. The hidden state of the RNN, in theory, summarizes all of the information from the past, so the hidden state in the DKT model can therefore be conceived as the latent knowledge state of a student that is resulted from his past learning trajectory. This latent knowledge state is then disseminated to the output layer to compute the output vector $\y_{{t}}$, which represents the probabilities of answering each KC correctly. If the student has a sequence of question-and-answer interactions of length $T$, the DKT model maps the inputs $(\x_1, \x_2, \dots, \x_{T})$ to the outputs $(\y_1, \y_2, \dots, \y_{T})$ accordingly. The unfolded RNN architecture for DKT is visualized in Figure~\ref{fig:rnn_architecture}, with a high-level interpretation.

However, since all of the information captured by the RNN lives in a same vector space in the hidden layer, the DKT model is consequently difficult to provide consistent prediction across time and therefore failing to pinpoint accurately which KCs a student is good at or unfamiliar with~\cite{LS2018_Yeung_DKTP}. Another novel neural network architecture, dynamic key-value memory network (DKVMN), is therefore proposed to alleviate this problem~\cite{WWW2017_Zhang_DKVMN}. 

DKVMN exploits the (single-head) attention mechanism in the memory augmented neural network~\cite{Nature2016_Graves_MANN} to model the skill acquisition process with two types of memory -- the static \textit{key memory} and the dynamic \textit{value memory}. The static \textit{key memory} and the dynamic \textit{value memory} are an analogue of the Python dictionary. The \textit{key memory} is immutable and stores the embedding vectors of different \textit{query keys} for the attention mechanism; while the \textit{value memory} is mutable and stores the \textit{numerical values} of the corresponding \textit{key}. To adapt the DKVMN model in the knowledge tracing setting, consider that there is a set of latent concepts (\textit{key memory}) which is associated with a set of corresponding latent knowledge states (\textit{value memory}). These latent concepts underlie all of the available KCs in a dataset, so each KC is a combination of these latent concepts. To construct a knowledge state of a KC, the same combination is applied on those corresponding latent knowledge states. This combination process generates vector representations of the KC and the knowledge state. The DKVMN model then estimates $P(a)$ based on the extracted KC and knowledge state vectors. In order to govern the KC acquisition process, the latent knowledge states in the \textit{value memory} are updated when a student has done a learning task. Zhang et al. shows that the DKVMN model not only outperforms the DKT model without suffering from overfitting, but also discovers underlying concepts for input KCs precisely.

Although DKVMN has provided an elegant methodology to model the KC acquisition process by distributing the vector representations of a KC and its knowledge state into different memory locations, these vector representations are still lack of interpretability, which is a common problem in most of the deep learning models. Therefore, there have been a tension in the learning science community between highly structured models, e.g., IRT, BKT and PFA, whose parameters have a direct psychological interpretation, versus, a highly complex but general-purpose models, e.g., DKT and DKVMN, whose parameters and representations are difficult to interpret. In the later section, we would like to relieve this tension by merging the IRT model and the DKVMN model so as to make the DKVMN model explainable.
\section{Deep Item Response Theory} \label{section:ModelFormulation}

\begin{figure*}[!t]
	\centering
    \includegraphics[width=0.9\linewidth]{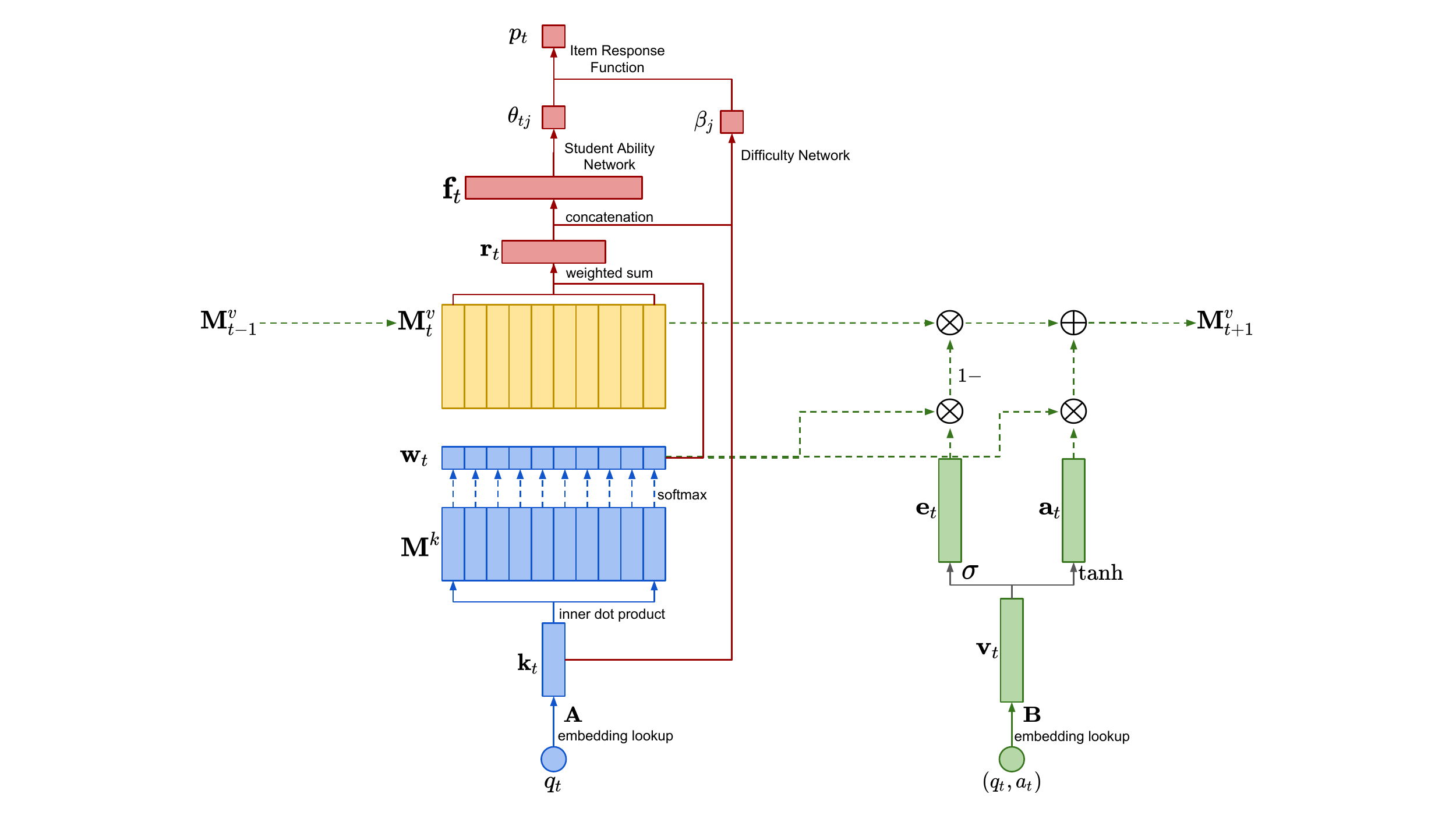}
    \caption{Network architecture for the Deep-IRT model. The model is drawn at time $t$ only. The blue components describe the process of getting the attention weight, the green components describe the process of updating the value memory and the red components describe the process of making a prediction. The $\otimes$ and $\oplus$ represent element-wise multiplication and addition, respectively.}
    \label{fig:deep_irt_architecture}
\end{figure*}

In this section, we explicate the Deep-IRT model which is a synthesis of the DKVMN model and the IRT model. Firstly, we explain the working mechanism of the DKVMN model mathematically. Then, we propose to augment the DKVMN model with the \emph{student ability network} and the \emph{difficulty network} so as to output the student ability level and the KC difficulty level that are required in the IRT model. Eventually, we use the one-parameter IRT model to predict the probability that a student will answer a KC correctly based on the estimated student ability and the KC difficulty level.

\subsection{Working Mechanism of DKVMN}
In summary, the DKVMN model works as follows: at time~$t$, it first receives a KC $q_t$, then predicts the probability of answering $q_t$ correctly, and eventually updates the memory using the question-and-answer interaction $(q_t, a_t)$. 

Assume that there are $N$ latent concepts underlying $Q$ distinct KCs. These latent concepts are stored in the key memory $\M^k \in \Rbb^{N \times d_k}$, where $d_k$ is the embedding size of a key memory slot. The student's knowledge states on these latent concepts are stored in the value memory $\M^v \in \Rbb^{N \times d_v}$, where $d_v$ is the embedding size of a value memory slot. To trace the student's knowledge state, the DKVMN model involves three major steps -- getting an attention weight, making a prediction and updating the value memory.

\subsubsection{Getting Attention Weight}
The DKVMN model can make a prediction when it receives an input $q_t$ at any time $t$. It first extracts the embedding vector of $q_t$ from a KC embedding matrix $\A \in \Rbb^{Q \times d_k}$. Then, it uses this embedding vector, denoted as $\k_t \in \Rbb^{d_k}$ later on, to query the key memory matrix $\M^k$ in the DKVMN model. The query result is the weighting of how much attention should be paid on each value memory slot. This attention weight $\w_t \in \Rbb^{N}$ is computed by the softmax activation of the inner product between $\k_t$ and each key memory slot $\M^k_i$:
\begin{align}
	w_{ti} &= \text{Softmax}(\M^k_i \k_t),
\end{align}
where $\sum_{i=1}^{N} w_{ti} = 1$ and $w_{ti}$ is the $i$-th element in the weight vector~$\w_t$, and $\M^k_i$ is the $i$-th row-vector of $\M^k$.

\subsubsection{Making Prediction}
With the attention weight~$\w_t$, the DKVMN model can predict the probability that a student will answer $q_t$ correctly by the following process. First, the DKVMN model reads the latent knowledge state in the value memory~$\M^v_t$ at time~$t$ to form a read vector 
\begin{align}
    \r_t = \sum_{i=1}^{N} w_{ti} (\M^v_{ti})^T,
\end{align}
where $\M^v_{ti}$ is the $i$-th row-vector of $\M^v_t$. Then, the read vector $\r_t$ and the KC embedding vector $\k_t$ are concatenated vertically and disseminated to a fully connected layer with the hyperbolic tangent activation so as to generate a feature vector $\f_t$. The feature vector $\f_t$ is then used to calculate the probability $p_t$ that the student will answer the KC $q_t$ correctly. These steps can be expressed mathematically as follows:
\begin{align} 
    \f_t &= \tanh (\W_f [\r_t, \k_t] + \b_f), \\
	p_t &= P(a_t) = \sigma(\W_p \f_t + \b_p),
\end{align}
where $[\cdot]$ denotes concatenation, and both the sigmoid function $\sigma(\cdot)$ and the hyperbolic tangent $\tanh(\cdot)$ are applied in an element-wise manner. These are parameterized by a weight matrix $\W$ and a bias vector $\b$ with appropriate dimensions. Here we adopt $p_t$ to represent $P(a_t)$ for the sake of avoiding confusion when we state the loss function later on. 

\subsubsection{Updating Value Memory}
The DKVMN model updates the value memory $\M^v_t$ based on the input tuple $(q_t, a_t)$ and the attention weight $\w_t$. The DKVMN model first retrieves an embedding vector of $(q_t, a_t)$ from a KC-response embedding matrix $\B \in \Rbb^{2Q \times d_v}$. This embedding vector, denoted as  $\v_t \in \Rbb^{d_v}$ later on, represents the knowledge growth after working on the KC $q_t$ with the correct label $a_t$. When updating the memory, some of the memory is first erased with an erase vector $\e_t \in \Rbb^{d_v}$ before new information is added to the memory with the add vector $\a_t \in \Rbb^{d_v}$. Erasing the memory offers the ability of forgetting similar to the LSTM cell. All in all, each value memory slot is updated as follows:
\begin{align}
	\e_t &= \sigma (\W_e \v_t + \b_e),\\
	\a_t &= \tanh (\W_a \v_t + \b_a),\\
	\tilde{\M}^v_{t+1, i} &= \M^v_{ti} \otimes (\mathbf{1} - w_{ti} \e_t)^T,	\\
	\M^v_{t+1, i} &= \tilde{\M}^v_{t+1, i} + w_{ti} \a_t^T,
\end{align}
where $\mathbf{1}$ is a vector of all ones, and $\otimes$ represents element-wise multiplication. 

\subsection{Student Ability and Difficulty Networks}

% it is for Section 4
%%%%%%%%%%%%%%%%%%%%%%%%%%%%%%%%%%%%%%%%%%%%%%%%%%%%%%%%%
%               Table: Dataset Summary                  %
%%%%%%%%%%%%%%%%%%%%%%%%%%%%%%%%%%%%%%%%%%%%%%%%%%%%%%%%%
\begin{table*}[t]
    \centering
    \caption{The summary of datasets.}
    \label{tab:dataset_summary}
    \begin{tabular}{l r r r r r r}
        \toprule
        Dataset & No. students & No. skills & No. questions & No. interactions & Sequence length & Correct rate \\
        \midrule
        ASSIST2009 & 4,151 & 110 & 26,684 & 325,637 & $78.45 \pm 155.86$ & 65.84\% \\
        ASSIST2015 & 19,840 & 100 & N/A & 683,801 & $34.47 \pm 41.39$ & 73.18\% \\
        Statics2011 & 333 & 156 & 1,223 & 189,297 & $568.46 \pm 370.30$ & 76.54\% \\
        Synthetic & 2000 & 5 & 50 & 100,000 & $50.00 \pm 0.00$ & 58.83\% \\
        XXXX-F1toF3 & 310 & 99 & 2266 & 51,283 & $165.43 \pm 163.65$ & 46.69\% \\ 
        \bottomrule
    \end{tabular}
\end{table*}

As the architecture of DKVMN is elegant, it can be easily augmented to provide other meaningful information during the model influence. Firstly, the knowledge state of each latent concept can be exploited to calculate the student ability. Specifically, when the DKVMN model receives a KC~$q_t$, it forms the feature vector~$\f_t$ during the influence. As $\f_t$ is the concatenation of the read vector~$\r_t$ and the KC embedding vector~$\k_t$, it contains information of both the student's knowledge state on $q_t$ and the embedded information of $q_t$. We believe that the $\f_t$ can be used to infer the student ability on $q_t$ by further processing $\f_t$ via a neural network. Similarly, the difficulty level of $q_t$ can be elicited by passing the KC embedding vector~$\k_t$ to a neural network. 

According to the purpose of the neural network, we call these two networks to be student ability network and difficulty network. Using a single fully-connected layer, we express these two networks as follows:
\begin{align}
    \theta_{tj} &= \tanh (\W_{\theta} \f_t + \b_{\theta}), \label{eq:deep-irt-theta} \\
    \beta_{j} &=  \tanh (\W_{\beta} \q_t + \b_{\beta}), \label{eq:deep-irt-beta}
\end{align}
where $\theta_{tj}$ and $\beta_{j}$ can interpreted as the student ability on the KC $j$ at time $t$ and the difficulty level of the KC $j$, respectively.
We use the hyperbolic tangent to be the activation function for both networks such that both outputs are scaled into the range $(-1, 1)$. Then, these two values are passed to the item response function to calculate the probability that a student will answer the KC $j$ correctly:
\begin{equation}
    p_t = \sigma (3.0*\theta_{tj} - \beta_{j}) \label{eq:deep-irt-output}.
\end{equation}
The output of the student ability network are multiplied by a factor of $3.0$ for a practical reason~\cite{SAP2014_Yang_IRTValidity}. For example, if we do not scale up the student ability, the maximum value that can be obtained is $\sigma(1 - (-1)) = \sigma(2) = 0.881$. 

The network architecture at time $t$ is plotted in Figure~\ref{fig:deep_irt_architecture}. It should be noted that the student ability network and the KC difficulty network can be applied to any types of neural network. For example, these two networks can be inserted to the DKT model, i.e., the RNN, surrounded by the hidden layer and the output layer. By formulating the knowledge tracing task with both the DKVMN model and the IRT model, we are getting the best from two worlds. The model benefits from the advance of deep learning techniques such that it captures features that are hard to be human-engineered. On the other hand, we empower the explainability by introducing a well-known psychometric model which can be easily understood by many people. This idea of merging deep learning models and psychometric models can be applied elsewhere apart from the knowledge tracing task, e.g., forgetting curve~\cite{JMP2011_Averell_ForgettingCurve}.

\section{Experiments}
\subsection{Datasets}
We employ four public datasets and one proprietary dataset in our experiment. For the public datasets, we used the processed data that provided by Zhang et al.~\cite{WWW2017_Zhang_DKVMN}. The information of these datasets is summarized in Table~\ref{tab:dataset_summary}.

\textbf{ASSIST2009} This dataset is provided by the ASSISTments online tutoring platform and has been used in several papers for the evaluation of knowledge tracing models. The dataset contains $4,151$ students with $325,637$ question-and-answering interactions from $26,688$ questions of $110$ skills. The average sequence length of a student is $78.45$ with a standard deviation of $155.86$. The correct rate of this dataset is $65.84\%$. As most of the literature adopts the skill tag when they conducted the experiment, we also adopt the skill tag as the input to a model in our experiment.

\textbf{ASSIST2015} This dataset contains $19,840$ student responses for $100$ skills with a total of $683,801$ question-and-answering interactions. Although it contains more interactions than ASSIST2009, the average number of records per skill and student is actually smaller due to a larger number of students. The average sequence length of a student is $34.47$ with a standard deviation of $41.39$. The correct rate of this dataset is $73.18\%$. As this dataset only provides the skill tag, we adopt the skill tag as the input to a model. 

\textbf{Statics2011} This dataset is obtained from an engineering statics course with $189,927$ interactions from $333$ students and $1,223$ question tags of $156$ skills. The average sequence length of a student is $568.46$ with a standard deviation of $370.30$. The correct rate of this dataset is $76.54\%$. Moreover, the question tag is used as the input to a model.

\textbf{Synthetic} Piech et al.~\cite{NIPS2015_Piech_DKT}
also simulated $2000$ virtual students' question-and-answer trajectories. Each student answers the same sequence of $50$ questions each of which belong to a single concept $j \in \{1,\ldots,5\}$ and has a difficulty level $\beta$, with an assigned ability $\theta$ of solving the task related to the concept $j$. The probability of a student answering a question correctly is defined based on the this IRT model $p(a) = c + \frac{1 - c}{1 + \exp(-(\theta - \beta))}$,
where $c$ denotes the probability of a student guessing it correctly and it is set to $0.25$. The question tag is used as the input to a model. 

\textbf{FSAI-F1toF3} This dataset is provided by the Find Solution Ai Limited and is collected via an adaptive learning tablet application called 4LittleTrees\footnote{More information about the 4LittleTrees can be found on \url{https://www.4littletrees.com/}.}. We extracted the student interactions that are related to mathematics curriculum from F.1 to F.3 in Hong Kong\footnote{F.1 (Form 1) means the first grade in the secondary school, which is equivalent to the 7th grade elsewhere. Thus, F.1 to F.3 is equivalent to 7th to 9th grade.}. It consists of $51,283$ question-and-answer interactions from $310$ students on $2,266$ questions of $99$ skills. The average sequence length of a student is $165.43$ with a standard deviation of $163.65$. The correct rate of this dataset is $46.69\%$. For this dataset, we use the question tag as the input to a model.

\subsection{Implementation}

%%%%%%%%%%%%%%%%%%%%%%%%%%%%%%%%%%%%%%%%%%%%%%%%%%%%%%%%%
%               Table: Experiment Result                %
%%%%%%%%%%%%%%%%%%%%%%%%%%%%%%%%%%%%%%%%%%%%%%%%%%%%%%%%%

\begin{table*}[t]
    \centering
    \caption{The average test results of the evaluation measures, as well as their standard deviations, from 5 trials are reported. As the PFA model can be learned by a closed-form solution, the learned parameters and thus its performance are the same in every trail. Therefore, the standard deviation is not reported for the PFA model.}
    \label{tab:result_model_performance}
    \begin{adjustbox}{max width=1.0\textwidth,center}
        \begin{tabular}{l | rrr | rrr | rrr | rrr}
            \toprule
            \multirow{2}{*}{Dataset} 
            & \multicolumn{3}{c|}{PFA} 
            & \multicolumn{3}{c|}{DKT} 
            & \multicolumn{3}{c|}{DKVMN} 
            & \multicolumn{3}{c}{Deep-IRT} 
            \\
            & AUC & Acc & Loss 
            & AUC & Acc & Loss 
            & AUC & Acc & Loss 
            & AUC & Acc & Loss
            \\
            \hline
            ASSIST2009
            & 59.68 & 69.24 & 7.08
            & 81.56 $\pm$ 0.18 & \textbf{77.17 $\pm$ 0.04} & \textbf{5.26 $\pm$ 0.01}
            & 81.61 $\pm$ 0.06 & 77.01 $\pm$ 0.04 & 5.29 $\pm$ 0.01 
            & \textbf{81.65 $\pm$ 0.02} & 77.00 $\pm$ 0.06 & 5.30 $\pm$ 0.01 
            \\
            ASSIST2015 
            & 52.85 & 73.37 & 6.13
            & 72.85 $\pm$ 0.05 & \textbf{75.29 $\pm$ 0.02} & \textbf{5.69 $\pm$ 0.01}
            & \textbf{72.94 $\pm$ 0.06} & 75.18 $\pm$ 0.03 & 5.71 $\pm$ 0.01 
            & 72.88 $\pm$ 0.07 & 75.14 $\pm$ 0.02 & 5.72 $\pm$ 0.01
            \\
            Statics2011 
            & 64.99 & 79.85 & 4.64
            & 82.71 $\pm$ 0.18 & 81.37 $\pm$ 0.08 & 4.29 $\pm$ 0.02 
            & \textbf{83.17 $\pm$ 0.11} & \textbf{81.57 $\pm$ 0.05} & \textbf{4.24 $\pm$ 0.01} 
            & 83.09 $\pm$ 0.12 & 81.56 $\pm$ 0.04 & 4.24 $\pm$ 0.01 
            \\
            Synthetic 
            & 61.68 & 65.20 & 8.01
            & 81.65 $\pm$ 0.06 & 74.84 $\pm$ 0.09 & 5.79 $\pm$ 0.02
            & 82.97 $\pm$ 0.06 & 75.58 $\pm$ 0.07 & 5.62 $\pm$ 0.02 
            & \textbf{82.98 $\pm$ 0.07} & \textbf{75.61 $\pm$ 0.04} & \textbf{5.61 $\pm$ 0.01}
            \\
            FSAI-F1toF3 
            & 54.52 & 54.57 & 10.46
            & \textbf{69.42 $\pm$ 0.34} & \textbf{64.11 $\pm$ 0.29} & \textbf{8.26 $\pm$ 0.07}
            & 68.40 $\pm$ 0.89 & 63.40 $\pm$ 0.15 & 8.42 $\pm$ 0.03 
            & 68.69 $\pm$ 0.28 & 63.43 $\pm$ 0.24 & 8.42 $\pm$ 0.06 
            \\
            \bottomrule
        \end{tabular}
    \end{adjustbox}
\end{table*}

\begin{table*}[t]
    \centering
    \caption{The corresponding hyperparameter settings of the models reported in Table~\ref{tab:result_model_performance}.}
    \label{tab:result_hyperparameter}
    \begin{adjustbox}{max width=1.0\textwidth,center}
        \begin{tabular}{l | rr | rrr | rrr}
            \toprule
            \multirow{2}{*}{Dataset} 
            & \multicolumn{2}{c|}{DKT} 
            & \multicolumn{3}{c|}{DKVMN} 
            & \multicolumn{3}{c}{Deep-IRT} 
            \\
            & s.~dim & \# params 
            & $N$ & s.~dim & \# params 
            & $N$ & s.~dim & \# params 
            \\ 
            \hline
            ASSIST2009 
            & 100 & $140,011$ 
            & 10 & 50 & $27,801$ 
            & 20 & 100 & $67,602$
            \\ 
            ASSIST2015 
            & 200 & $341,901$ 
            & 100 & 50 & $35,301$
            & 20 & 50 & $27,352$ 
            \\ 
            Statics2011 
            & 100 & $1,142,824$
            & 5 & 200 & $836,701$ 
            & 5 & 100 & $398,502$
            \\ 
            Synthetic 
            & 200 & $251,851$ 
            & 50 & 100 & $55,501$ 
            & 20 & 50 & $19,852$
            \\ 
            FSAI-F1toF3 
            & 100 & $2,082,567$ 
            & 10 & 100 & $712,301$ 
            & 5 & 50 & $350,752$
            \\ 
            \bottomrule
        \end{tabular}
    \end{adjustbox}
\end{table*}

\subsubsection{Experiments Setting} 
We feed the input $q_t$ and $(q_t, a_t)$ to the network using their ID tag, where $ID(q_t) = q_t \in \{1,2,\ldots,Q\}$ and $ID(q_t, a_t) = q_t + a_t * Q \in \{1,2,\ldots,2Q\}$ if there are $Q$ distinct KCs. The IDs of $q_t$ and $(q_t, a_t)$ are used to lookup the embedding vectors in the the KC embedding matrix $\A$ and the KC-response embedding matrix $\B$, respectively. 

Prior to training, the KC embedding matrix $\A$ and the KC-response embedding matrix $\B$, as well as the key memory matrix $\M^k$, the value memory matrix $\M^v$ and other model parameters $\W$ and $\b$, are initialized randomly from a Gaussian distribution with zero mean and a small standard deviation. All of these model parameters are learned during the training process by minimizing the cross-entropy loss: 
\begin{equation}
    \Lcal = - \sum_t \left( a_t \log p_t + (1 - a_t) \log (1-p_t) \right).
\end{equation} 
We learn the model using the Adam optimization with a learning rate of $0.003$ and a batch size of $32$. We also consistently set the norm clipping threshold to $10.0$ in order to avoid gradient exploding. In addition, since all of the input sequences are of different lengths, all sequences, except the Synthetic dataset, are set to be a length of $200$. For sequences that have less than $200$ time-steps, they are padded with zeros to fill up the remaining time steps. Accordingly, masking is also applied when computing the loss. 

We implement the models using Tensorflow on a computer with a single NVIDIA RTX 2080 GPU.

\subsubsection{Hyperparameter Selection and Evaluation}
Of all the datasets, $30\%$ of the sequences are held out as a test set and the remaining $70\%$ are used as a training set. Furthermore, five-fold cross-validation is applied on the training set to perform hyperparameter selection. The network architectures of the DKT, DKVMN and Deep-IRT models are varied with different numbers of state dimension and memory size. A grid search is applied over the combinations of state dimension and memory size.

As for the DKT model, the hidden layer size is chosen from \{10, 50, 100, 200\}. For the DKVMN and Deep-IRT models, the dimensions of key memory matrix ($d_k$) and the value memory matrix ($d_v$) are chosen from \{10, 50, 100, 200\} as well. We also choose the memory dimension $N$ for the DKVMN and Deep-IRT models from \{5, 10, 20, 50, 100\}. As the search space is large if we perform a grid search over all possible combinations of $N$, $d_k$ and $d_v$, we therefore set $d_k = d_v$ for simplicity. Lastly, we use the combination that results in the smallest cross-entropy loss to retrain the model with the entire training set and evaluate the model performance on the test set.

To report the model performance, we run the training and evaluation process for 5 times. The average and standard deviation of area under the ROC curve (AUC), the accuracy and the cross-entropy loss are reported. AUC provides a robust metric for binary prediction evaluation. When interpreting the value of AUC, the larger the AUC score is, the better the prediction performance is of the model. An AUC score of 0.5 indicates that the model performance is merely as good as random guess.

\subsection{Results}
The model performance of the experiment is shown in Table~\ref{tab:result_model_performance} and the corresponding hyperparameters are stated in Table~\ref{tab:result_hyperparameter}. In addition, we also include the performance of the PFA model in Table~\ref{tab:result_model_performance} as a baseline model for reference. It is noted that the model performance of the DKT model in our experiment is better than the one reported in~\cite{WWW2017_Zhang_DKVMN}, even thought we adopt the same processed data provided by Zhang et al.~\cite{WWW2017_Zhang_DKVMN}. This might be attributed to different random seeds and deep learning libraries that are used to train the model.

For the ASSIST2009 and ASSIST2015 datasets, the DKT model achieves the highest accuracy and the smallest loss, while the Deep-IRT model and the DKVMN model result in the highest AUC score in the ASSIST2009 and the ASSIST2015, respectively. As for the Statics2011 dataset, the DKVMN model obtains the highest AUC score, the highest accuracy and the smallest loss. Regarding to the Synthetic dataset, the Deep-IRT results in the best performance in AUC, accuracy and loss. On the FSAI-F1toF3 dataset, the DKT model performs the best. All in all, the DKT, DKVMN and Deep-IRT models have a similar performance.

In addition, we conduct two-tailed independent t-tests on each dataset and each evaluation measure between the Deep-IRT model and the DKVMN model. We found that the difference between their performance is not significant for majority of the datasets. The p-values of the t-tests are reported as follows in Table~\ref{tab:experiment_p_value}:
\begin{table}[h]
    \centering
    \caption{The p-values of the independent t-tests between the DKVMN model and the Deep-IRT model.}
    \label{tab:experiment_p_value}
    \begin{tabular}{l r r r}
        \toprule
        \multirow{2}{*}{Dataset} 
        & \multicolumn{3}{c}{p-value}
        \\
        & AUC & Accuracy & Loss \\
        \midrule
        ASSIST2009 
        & 0.2581 & 0.7061 & 0.7061 \\
        ASSIST2015
        & 0.1695 & 0.0472 & 0.0472 \\
        Statics2011
        & 0.3239 & 0.7836 & 0.7836 \\
        Synthetic
        & 0.7565 & 0.3907 & 0.3907 \\
        FSAI-F1toF3
        & 0.5119 & 0.8101 & 0.8101 \\
        \bottomrule
    \end{tabular}
\end{table}

Although we cannot claim that their performance is, more or less, the same based on the large p-value, this result, however, might imply that the Deep-IRT model potentially retains the performance of the DKVMN model.

\begin{figure*}[t]
    \centering
    \includegraphics[width=0.79\linewidth]{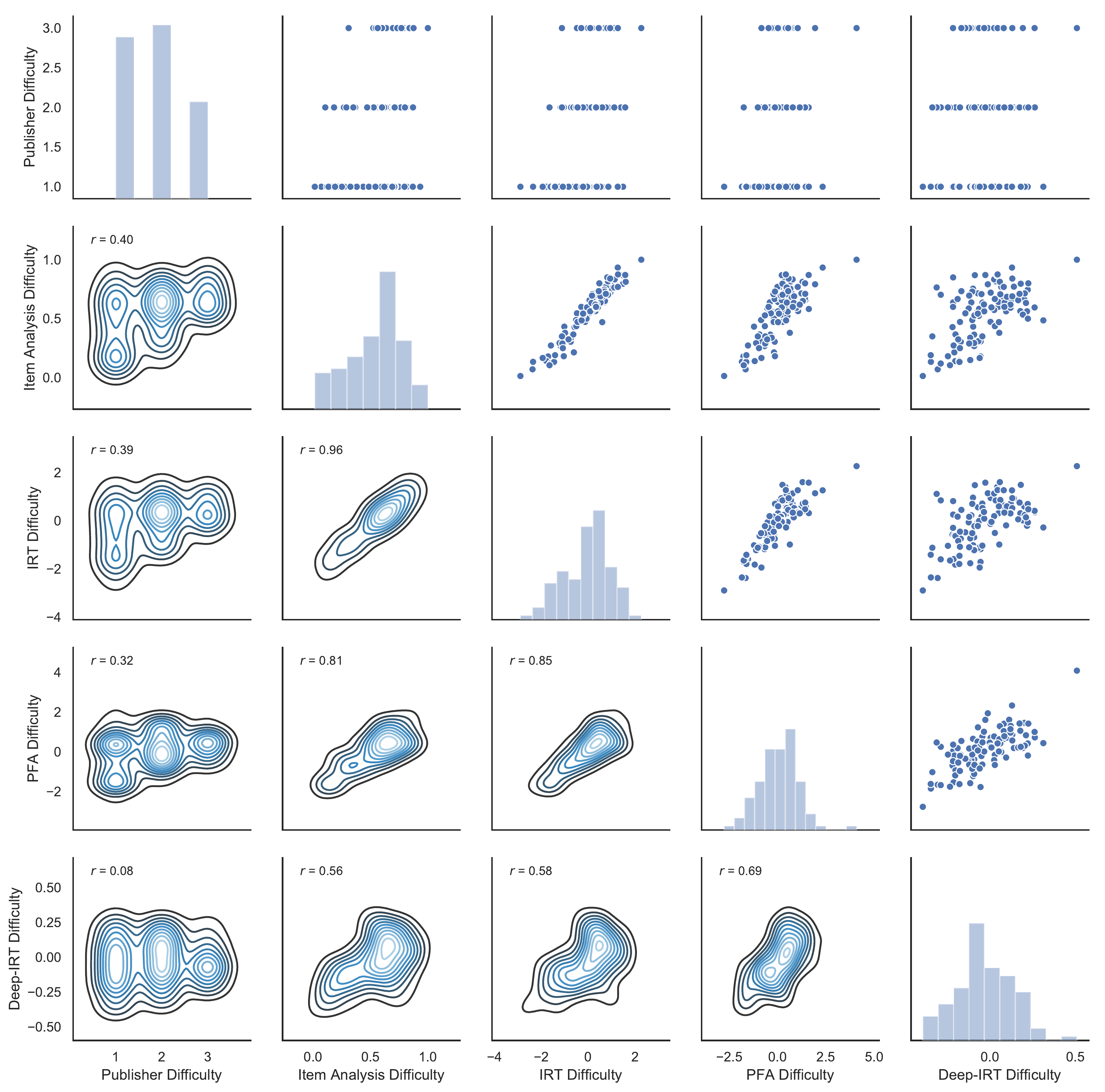}
    \caption{Pairwise comparison of the difficulty level obtained from different sources. The positions in the pairs plot are ordered according to the evaluation setting and the complexity of getting the difficulty level.}
    \label{fig:difficulty_pairwise}
\end{figure*}

\section{Discussion}
\begin{figure*}
    \centering
    \includegraphics[width=\linewidth]{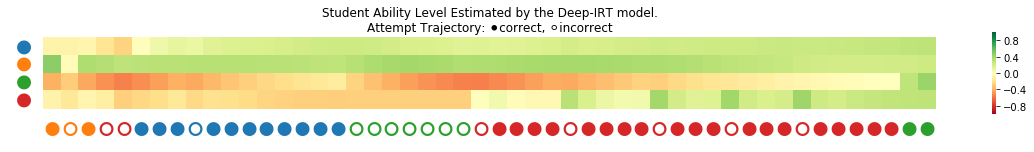}
    \includegraphics[width=\linewidth]{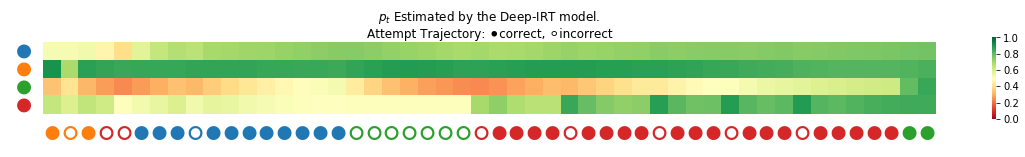}
    \caption{An example of a student's learning trajectory from the ASSIST2009 dataset. The student ability (top) and the probability prediction (bottom) are shown. The labels in the vertical dimension correspond to different skill tags. They are, from top to bottom, ``equation solving two or fewer steps''~(blue), ``ordering factions''~(orange), ``finding percents''~(green) and ``equation solving more than two steps''~(red), respectively. The learning trajectory is shown in the horizontal dimension by filled or hollow circles with corresponding colors. The filled and the hollow circles represent a correct response and an incorrect response, respectively.}
    \label{fig:assist2009_student_ability}
\end{figure*}

\subsection{Going Deeper in Difficulty Level}
To evaluate the KC difficulty estimated from the Deep-IRT model, we compare the difficulty level learned for the FSAI-F1toF3 dataset with four other sources. The reason why we use the proprietary dataset is that we have the individual questions' difficulty level provided by the publisher. Each question is associated with a difficulty level in $\{1,2,3\}$ which represents easy, medium and hard, respectively. 

The second source of difficulty level is calculated according to the item analysis~\cite{Washington_ItemAnalysis}. The difficulty level of the item analysis is the percentage of students who answer a question correctly in a test environment. Yet, to be consistent with interpretation with other models, we adopt the percentage of students who answer a question \emph{incorrectly}. Furthermore, as our dataset is not collected from a test environment, a student can answer a same question multiple times. Thus, we only adopt the student's first attempt when calculating the difficulty level. Moreover, we only consider the question on which at least 10 distinct students has answered. 

The third source of difficulty level is learned by the one-parameter IRT model. We also use the student's first attempt, only, on a question to learn the IRT model for the sake of avoiding multiple attempts on the same question. Moreover, we adopt $\sigma (\theta_{i} - \beta_{j})$ to be the item response function when learning the IRT model.

Lastly, we trained a PFA model to extract the difficulty level in a knowledge tracing setting, rather than a test environment setting. In other words, we use the entire student's learning trajectory to learn the question's difficulty level.

Since there are more than two thousands questions in this dataset, we only evaluate a set of questions that belong to a subset of skills. This subset contains five skills that constitute around a fifth of the interactions in the FSAI-F1toF3 dataset, and  has in total $131$ questions. These skills are ``Significant Figures'', ``Approximation and Errors in Measurement'', ``Index Notation'', ``Laws of Indices'' and ``Polynomials''. We visualize the difficulty level obtained from different sources in a pairs plot in Figure~\ref{fig:difficulty_pairwise} with the Pearson correlation $r$ stated in the lower triangular part of the pairs plot. The positions in the pairs plot are ordered according to the evaluation setting and the complexity of getting the difficulty level.

The pairs plot reveals that the difficulty level learned from the Deep-IRT model aligns with most of the other sources with a strong correlation, except for the difficulty level provided by the publisher. Moreover, it is observed that the more similar the models' evaluation setting and complexity are, the higher the Pearson correlation is between the models. For example, the Pearson correlation between the difficulty level from the item analysis model and the IRT model is 0.96, while the Pearson correlation between the item analysis model and the Deep-IRT model is 0.56. Furthermore, it is observed that the difficulty level provided by the publisher is moderately correlated to the one obtained from the item analysis model (0.40) and the IRT model (0.39), but weakly correlated to the one obtained from the Deep-IRT model (0.08). Thus, it would be interesting to examine whether the difficulty level inferred from the Deep-IRT model would be more accurate than other traditional models.

\subsection{Going Deeper in Student Ability}
As stated in~\cite{LS2018_Yeung_DKTP}, there are two problems exist in the DKT model. The first one is that the DKT model fails to reconstruct the observed input. This means that the estimated performance of a student decreases even if the student makes a successful attempt, and vice versa. The second problem is that the estimated performance of different KCs is not consistent over time. This means that the student mastery level alternates between mastered and not-yet-mastered during the model influence. These two behaviors are undesirable and therefore we would like to examine if these problems exist in the Deep-IRT model.

We randomly select a student in the ASSIST2009 dataset and then evaluate the transition of the student's ability level and the prediction probability of getting the next KC correctly in the next time-step during the learning trajectory by visualizing the first 50 attempts of that student (shown in Figure~\ref{fig:assist2009_student_ability}). In the first 50 attempts, the student has attempted the skills of ``equation solving two or fewer steps''~(shown in blue), ``ordering factions''~(orange), ``finding percents''~(green) and ``equation solving more than two steps''~(red). 

As observed from the figure, the prediction transition is smooth while there are some counter intuitive prediction. For example, when the student answers ``equation solving more than two steps''~(red) incorrectly, the corresponding estimated student ability level and the prediction probability increase, instead. Moreover, it seems that the model cannot identify the prerequisite relationship among the skills. For instance, when the student is answering the skill ``equation  solving  two  or  fewer  steps'' (blue) correctly, the estimated student ability level and the prediction probability of the skill ``equation solving more than two steps''~(red) decrease. However, it is expected if the student is doing well in the skill ``equation  solving  two  or  fewer  steps'' (blue), it should, at least, does not have an adverse effect on the skill ``equation solving more than two steps''~(red). Thus, it would be interesting to investigate methods to make deep learning based knowledge tracing model reliable, e.g., the prediction-consistent regularization proposed in~\cite{LS2018_Yeung_DKTP}.

\section{Conclusion}
In this paper, we propose the Deep-IRT model which empowers the deep learning based knowledge tracing model with explaniability. Experiments show that the Deep-IRT model retains the performance of the deep learning based knowledge tracing model while simultaneously being able to estimate the KC difficulty level and the student ability level over time. Moreover, the difficulty level estimated by the Deep-IRT model aligns with the difficulty level obtained by other traditional methods, e.g., the IRT model and the item analysis. Thus, it potentially provides an alternative way to estimate KC's difficulty level by utilizing the entire learning trajectory, rather than the traditional educational testing environment. However, there are still rooms for improvement on learning the inherent KC dependency. One of the potential ways is to learn a better KC vector representation by incorporating the content of questions~\cite{AAAI2018_Su_Exercise}.

\bibliographystyle{plainnat}
\balance
\bibliography{references}

\end{document}